\documentclass[11pt,a4paper]{article}
\usepackage[hyperref]{eacl2021}
\usepackage{times}
\usepackage{latexsym}

\usepackage{microtype}

\usepackage[utf8]{inputenc}
\usepackage{subcaption}
\usepackage{graphicx}

\usepackage[]{todonotes}
\usepackage{booktabs}
\usepackage{multirow}
\usepackage{covington}
\usepackage{svg}
\usepackage{enumerate}
\usepackage{bm}

\usepackage{dblfloatfix} 

\usepackage{makecell}
\usepackage{mathtools}
\usepackage{adjustbox}

\aclfinalcopy 
\def\aclpaperid{68}

\title{Explaining and Improving BERT Performance\\on Lexical Semantic Change Detection}

\author{Severin Laicher, Sinan Kurtyigit, Dominik Schlechtweg,\\\bf Jonas Kuhn and Sabine Schulte im Walde\\
Institute for Natural Language Processing, University of Stuttgart\\
\small
{\tt \{laichesn,kurtyisn,schlecdk,jonas,schulte\}@ims.uni-stuttgart.de}}

\date{}

\begin{document}
\maketitle
\begin{abstract}
Type- and token-based embedding architectures are still competing in lexical semantic change detection. The recent success of type-based models in SemEval-2020 Task 1 has raised the question why the success of token-based models on a variety of other NLP tasks does not translate to our field. We investigate the influence of a range of variables on clusterings of BERT vectors and show that its low performance is largely due to orthographic information on the target word, which is encoded even in the higher layers of BERT representations. By reducing the influence of orthography we considerably improve BERT's performance.
\end{abstract}

\section{Introduction}
Lexical Semantic Change (LSC) Detection has drawn increasing attention in the past years \citep{kutuzov-etal-2018-diachronic,2018arXiv181106278T,hengchen2021challenges}. Recently, SemEval-2020 Task 1 and the Italian follow-up task DIACR-Ita provided a multi-lingual evaluation framework to compare the variety of proposed model architectures \citep{schlechtweg-etal-2020-semeval,diacrita_evalita2020}. Both tasks demonstrated that type-based embeddings outperform token-based embeddings. This is surprising given that contextualised token-based approaches have achieved significant improvements over the static type-based approaches in several NLP tasks over the past years \citep{peters-etal-2018-deep,devlin-etal-2019-bert}. 

In this study, we relate model results on LSC detection to results on the word sense disambiguation data set underlying SemEval-2020 Task 1. This allows us to test the performance of different methods more rigorously, and to thoroughly analyze results of clustering-based methods.
We investigate the influence of a range of variables on clusterings of BERT vectors and show that its low performance is largely due to orthographic information on the target word which is encoded even in the higher layers of BERT representations. By reducing the influence of orthography on the target word while keeping the rest of the input in its natural form we considerably improve BERT's performance.

\section{Related work}

Traditional approaches for LSC detection are type-based \citep{Dubossarskyetal19,Schlechtwegetal19}. This means that not every word occurrence is considered individually (token-based); instead, a general vector representation that summarizes every occurrence of a word (including polysemous words) is created. The results of SemEval-2020 Task 1 and DIACR-Ita \citep{diacrita_evalita2020,schlechtweg-etal-2020-semeval} demonstrated that overall type-based approaches \citep{asgari-etal-2020-emblexchange,kaiser-etal-2020-roots,prazak-etal-2020-uwb} achieved better results than token-based approaches \citep{beck-2020-diasense,kutuzov-giulianelli-2020-uiouva,laicher-etal-2020-volente}. This is surprising, however, for two main reasons: (i) contextualized token-based approaches have significantly outperformed static type-based approaches in several NLP tasks over the past years  \citep{ethayarajh2019contextual}. (ii) SemEval-2020 Task 1 and DIACR-Ita both include a subtask on binary change detection that requires to discover small sets of contextualized usages with the same sense. Type-based embeddings do not infer usage-based (or token-based) representations and are therefore not expected to be able to find such sets \citep{schlechtweg-etal-2020-semeval}. Yet, they show better performance on binary change detection than clusterings of token-based embeddings \citep{kutuzov-giulianelli-2020-uiouva}.

\section{Data and evaluation}
We utilize the annotated English, German and Swedish datasets (ENG, GER, SWE) underlying SemEval-2020 Task 1 \citep{schlechtweg-etal-2020-semeval}. Each dataset contains a list of target words and a set of usages per target word from two time periods, $t_1$ and $t_2$ \citep{Schlechtweg2021dwug}. For each target word, a Word Usage Graph (WUG) was annotated, where nodes represent word usages, and weights on edges represent the (median) semantic relatedness judgment of a pair of usages, as exemplified in (\ref{ex:1}) and (\ref{ex:2}) for the target word \textit{plane}.
\begin{example}\label{ex:1}
Von Hassel replied that he had such faith in the \textbf{plane} that he had no hesitation about allowing his only son to become a Starfighter pilot.
\end{example}%
\begin{example}\label{ex:2}
This point, where the rays pass through the perspective \textbf{plane}, is called the seat of their representation.
\end{example}
The final WUGs were clustered with a variation of correlation clustering \citep{Bansal04}
(see Figure \ref{fig:graph1} in Appendix \ref{sec:wugs}, left) and split into two subgraphs representing nodes from $t_1$ and $t_2$ respectively (middle and right). Clusters are interpreted as senses, and changes in clusters over time are interpreted as lexical semantic change. \citeauthor{schlechtweg-etal-2020-semeval} then infer a binary change value $B(w)$ for Subtask 1 and a graded change value $G(w)$ for Subtask 2 from the two resulting time-specific clusterings for each target word $w$.

The evaluation of the shared task participants only relied on the change values derived from the annotation, while the annotated usages were not released. We gained access to the data set, which enables us to relate performances in change detection to the underlying data.\footnote{We had no access to the Latin annotated data. For the ENG clustering experiments we use the full annotated resource containing three additional graphs \citep{Schlechtweg2021dwug}.} We can also analyze the inferred clusterings with respect to bias factors, and compare their influence on inferred vs. gold clusterings. A further advantage of having access to the underlying data is that it reflects more accurately the annotated change scores. In SemEval-2020 Task 1 the annotated usages were mixed with additional usages to create the training corpora for the shared task, possibly introducing noise on the derived change scores.

\section{Models and Measures}
\label{sec:models}
\paragraph{BERT} 
Bidirectional Encoder Representations from Transformers \citep[BERT,][]{devlin-etal-2019-bert} is a transformer-based neural language model designed to find contextualised representations for text by analysing left and right contexts. The base version processes text in 12 different layers. In each layer, a contextualized token vector representation is created for every word. A layer, or a combination of multiple layers (we use the average), serves as a representation for a token. For every target word, we feed the usages from the SemEval data set into BERT and use the respective pre-trained cased base model to create token embeddings.\footnote{We first clean the GER usages by replacing historical with modern characters.}

\paragraph{Clustering}
LSC can be detected by clustering the token vectors from $t_1$ and $t_2$ into sets of usages with similar meanings, and then comparing these clusters over time \citep[cf.][]{Schutze1998,Navigli09}. This section introduces the clustering algorithms and clustering performance measures that we used. \textbf{Agglomerative Clustering} (AGL) is a hierarchical clustering algorithm starting with each element in an individual cluster. It then repeatedly merges those two clusters whose merging maximizes a predefined criterion. We use Ward's method, where clusters with the lowest loss of information are merged \citep{ward1963hierarchical}. Following \citet{giulianelli-etal-2020-analysing} and \citet{Martinc2020evolution}, we estimate the number of clusters $k$ with the \textbf{Silhouette Method} \citep{rousseeuw1987silhouettes}: we perform a cluster analysis for each $2 \leq k \leq 10$ and calculate the silhouette index for each $k$. The number of clusters with the largest index is used for the final clustering. 
The \textbf{Jensen-Shannon Distance} (JSD) measures the difference between two probability distributions \citep{Lin1991,DonosoS17}. We convert two time specific clusterings into probability distribution $P$ and $Q$ and measure their distance $JSD(P,Q)$ to obtain graded change values \citep{giulianelli-etal-2020-analysing,kutuzov-giulianelli-2020-uiouva}. If $P$ and $Q$ are very similar, the JSD returns a value close to 0. If the distributions are very different, the JSD returns a value close to 1.
\textbf{Spearman's Rank-Order Correlation Coefficient} $\rho$ measures the strength and the direction of the relationship between two variables \citep{bolboaca2006pearson} by correlating the rank order of two variables. Its values range from -1 to 1, where 1 denotes a perfect positive relationship between the two variables, and -1 a perfect negative relationship. 0 means that the two variables are not related.  

\paragraph{Cluster bias}
We perform a detailed analysis on what the inferred clusters actually reflect. We test hypotheses on \textit{word form}, \textit{sentence position}, \textit{number of proper names} and \textit{corpus}. The influence strength of each of these variables on the clusters is measured by the \textbf{Adjusted Rand Index} (ARI) \citep{Hubert/Arabie:85}
between the inferred cluster labels for each test sentence and a labeling for each test sentence derived from the respective variable. For the variable \textit{word form}, we assign the same label to each use where the target word has the same orthographic form (same string). If ARI = 1, then the inferred clusters contain only sentences where the target word has the same form.
For \textit{sentence position} each sentence receives label 0, if the target word is one of the first three words of the sentence, 2, if the target word is one of the last three words, else 1.\footnote{We assume that especially the beginning and ending of a sentence have a strong influence.}
For \textit{proper names} a sentence receives label 0, if no proper names are in the sentence, 1, if one proper name occurs, else 2.\footnote{The influence of proper names is only measured for ENG, since no POS-tagged data was readily available for GER.} The hypothesis that proper names may influence the clustering was suggested in \citet{martinc-etal-2020-context}. For \textit{corpora}, a sentence is labeled 0, if it occurs in the first target corpus, else 1. 

\paragraph{Average measures}
Given two sets of token vectors $V_1$ and $V_2$ from $t_1$ and $t_2$, \textbf{Average Pairwise Distance} (APD) is calculated by randomly picking $n$ vectors from both sets, calculating their pairwise cosine distances $d(x,y)$ where $x \in V_1$ and $y \in V_2$ and averaging over these.
\citep{Schlechtwegetal18,giulianelli-etal-2020-analysing}. We determine $n$ as the minimum size of $V_1$ and $V_2$. 
\textbf{APD-OLD/NEW} measure the average of pairwise distances within $V_1$ and $V_2$, respectively. They are calculated as the average distance of max. $10,000$ randomly sampled unique combinations of vectors from either $V_1$ or $V_2$. \textbf{COS} is calculated as the cosine distance of the respective mean vectors for $V_1$ and $V_2$ \citep{kutuzov-giulianelli-2020-uiouva}. 

\begin{table*}[ht]
\center
\begin{adjustbox}{width=0.43\textwidth}
\begin{tabular}{ l | l | c c c }
\toprule
& \textbf{Layer} & \textbf{Token} & \textbf{Lemma} & \textbf{TokLem} \\
\midrule
\multirow{3}{*}{\rotatebox[origin=c]{0}{\textbf{$\rho$}}}
& 1 & -.141 & -.033 & .100\\
& 12 & .205 & .154 & .168 \\
& 9-12 & .325 & \textbf{.345} & .293 \\
\midrule
\multirow{3}{*}{\rotatebox[origin=c]{0}{\textbf{ARI}}}
& 1 & .022 & .041 & .045\\
& 12 & .116 & .111 & .158 \\
& 9-12 & .150 & .159 & \textbf{.163} \\
\midrule
\multicolumn{4}{c}{}  \\
\midrule
\multirow{3}{*}{\rotatebox[origin=c]{0}{\textbf{Form}}}
& 1 & \textbf{.907} & .014 & .014\\
& 12 & \textbf{.389} & \textbf{.018} & \textbf{.077} \\
& 9-12 & \textbf{.334} & \textbf{.018} & \textbf{.051} \\
\midrule
\multirow{3}{*}{\rotatebox[origin=c]{0}{\textbf{Position}}}
& 1 & .001 & \textbf{.026} & \textbf{.024}\\
& 12 & \textbf{.012} & \textbf{.012} & \textbf{.015} \\
& 9-12 & .002 & \textbf{.007} & \textbf{.003} \\
\midrule
\multirow{3}{*}{\rotatebox[origin=c]{0}{\textbf{Corpora}}}
& 1 & \textbf{.019} & \textbf{.021} & \textbf{.033}\\
& 12 & \textbf{.078} & \textbf{.056} & \textbf{.082} \\
& 9-12 & \textbf{.056} & \textbf{.044} & \textbf{.063} \\
\midrule
\multirow{3}{*}{\rotatebox[origin=c]{0}{\textbf{Names}}}
& 1 & -.007 & .010 & .010\\
& 12 & \textbf{.025} & \textbf{.027} & \textbf{.033} \\
& 9-12 & .019 & \textbf{.022} & \textbf{.026} \\

\bottomrule
\end{tabular}
\end{adjustbox}
\quad
\begin{adjustbox}{width=0.43\textwidth}
\begin{tabular}{ l | l | c c c }
\toprule
& \textbf{Layer} & \textbf{Token} & \textbf{Lemma} & \textbf{TokLem} \\
\midrule
\multirow{3}{*}{\rotatebox[origin=c]{0}{\textbf{$\rho$}}}
& 1              &      -.265      &      -.062 &      -.170\\
& 12           &      .123      &      .427 &      \textbf{.624}\\
& 9-12   &      .122      &      .420 &      .533\\
\midrule
\multirow{3}{*}{\rotatebox[origin=c]{0}{\textbf{ARI}}}
& 1              &      .033      &      .002 &      .003\\
& 12           &      .119      &      .159 &      \textbf{.161}\\
& 9-12   &      .155      &      .142 &      .154\\
\midrule
\multicolumn{4}{c}{}  \\
\midrule
\multirow{3}{*}{\rotatebox[origin=c]{0}{\textbf{Form}}}
& 1              &      \textbf{.706}      &      .024 &      .004\\
& 12           &      \textbf{.439}     &      \textbf{.056} &      \textbf{.150}\\
& 9-12   &      \textbf{.420}      &      \textbf{.047} &      \textbf{.094}\\
\midrule
\multirow{3}{*}{\rotatebox[origin=c]{0}{\textbf{Position}}}
& 1              &      .005      &      \textbf{.023} &      \textbf{.027}\\
& 12           &      -.002      &      .005 &      -.002\\
& 9-12   &      \textbf{.009}      &      \textbf{.018} &      \textbf{.012}\\
\midrule
\multirow{3}{*}{\rotatebox[origin=c]{0}{\textbf{Corpora}}}
& 1              &      .074      &      .003 &      .005\\
& 12           &      \textbf{.110}      &      \textbf{.095} &      \textbf{.096}\\
& 9-12   &      \textbf{.107}     &      .068 &      \textbf{.089}\\
\midrule
\multirow{3}{*}{\rotatebox[origin=c]{0}{\textbf{Names}}}
& 1              &      -      &      - &      -\\
& 12           &      -      &      - &      -\\
& 9-12           &      -      &      - &      -\\

\bottomrule
\end{tabular}
\end{adjustbox}
\caption{Overview of English clustering scores  (left) and German clustering scores (right). Bold font indicates best scores for $\rho$ and ARI (top) or scores above all corresponding baselines for influence variables (bottom).}
\label{tbl: ClusterInfluence short}
\end{table*}
\section{Results}

\subsection{Clustering}
\label{clustering}

Because of the high computational load, we apply the clustering only to the ENG and the GER part of the SemEval data set. For this, we use BERT to create token vectors and cluster them with AGL, as described above. We then perform a detailed analysis of what the clusters reflect.\footnote{We also run most of our experiments with k-means \citep{Forgy:65}. Both algorithms performed similarly with a slight advantage for AGL. We therefore only report the results achieved using AGL.}

We report a subset of the clustering experiment results in Table \ref{tbl: ClusterInfluence short}, the complete results are provided in Appendix \ref{sec:performance}.
Table \ref{tbl: ClusterInfluence short} shows JSD performance on graded change ($\rho$), clustering performance (ARI) as well as the ARI scores for the influence factors introduced above, across BERT layers. For each influence factor we add two baselines: (i) The random baseline measures the ARI score of the influence factor using random cluster labels, and (ii) the actual baseline measures the ARI score between the true cluster labels and the influence factor. In other words, (i) and (ii) respectively answer the question of how strong the influence factor is by chance, and how strong it is according to the human annotation. The values of the two baselines are crucial: If an influence factor has an ARI score greater than both baselines, the clustering reflects the influence factor more than expected. If additionally the influence factor has an ARI score greater than the actual performance ARI score, the clustering reflects the partitioning according to the influence factor more than the clustering derived from human annotations. 

\paragraph{Word form bias} As explained above, the word form influence measures how strongly the inferred clusterings represent the orthographic forms of the target word. Table \ref{tbl: ClusterInfluence short} shows that for both GER and ENG the form bias of the raw token vectors (column `Token') is extremely high and always yields the highest influence score for each layer combination of BERT. Additionally, the influence of the word form is significantly higher when using lower layers of BERT. This fits well with the observations of \citet{jawahar2019does} that the lower layers of BERT capture surface features, the middle layers capture syntactic features and the higher layers capture semantic features of the text. With the first layer of BERT the sentences are almost exclusively ($.9$) clustered according to the form of the target word (e.g. plural/singular division). Even in the higher layers word form influence is considerable in both languages (layer 12: $\approx .4$). This strongly overlays the semantic information encoded in the vectors, as we can see in the low $\rho$ and ARI scores, which are negatively correlated with word form influence.\footnote{Note that it is very difficult to reach high ARI scores because ARI incorporates chance.}

The word form bias seems to be lower in GER than in ENG (layer 1: $.7$ vs. $.9$). However, this is misleading, as our approach to measure word form influence does not capture cases where vectors cluster according to subword forms as in the case of \textit{Ackergerät}. Its word forms differ as to whether they are written with an `h' or not, as in \textit{Ackergerät} vs. \textit{Ackergeräth}. As a manual inspection shows this is strongly reflected in the inferred clustering. However, these forms then further subdivide into inflected forms such as \textit{Ackergeräthe} and \textit{Ackergeräthes}, which is reflected in our influence variable. For these cases, our approach tends to underestimate the influence of the variable.

In order to reduce the influence of word form we experiment with two pre-processing approaches: (i) We feed BERT with lemmatised sentences (Lemma) instead of raw ones. (ii) We only replace the target word in every sentence with its lemma (TokLem). TokLem is motivated by the fact that BERT is trained on raw text. Thus, we assume that BERT is more familiar with non-lemmatised sentences and therefore expect it to work better on raw text. In order to continue working with non-lemmatised sentences we only remove the target word form bias by exchanging the target word with its lemma.   

As we can see in Table \ref{tbl: ClusterInfluence short}, lemmatisation strongly reduces the influence of word form, as expected.\footnote{In some cases it is however above the baselines, indicating that word form is correlated with other sentence features.} Accordingly, $\rho$ and ARI improve. However, it also leads to deterioration in some cases. Also, TokLem reduces the influence of word form and in most cases yields the overall maximum performance. The ARI scores for both languages are similar ($\approx .160$) while the $\rho$ performance varies very strongly between languages, achieving a very high score for GER ($.624$).  

Replacing the target word by its lemma form seems to shift the word form influence in the different layers: Especially for GER, layers 1 and 1+12 show the highest influences ($.706$ and $.687$) with Token (see also Appendix \ref{sec:performance}). In combination with TokLem, both layers are influenced the least ($.004$ and $.046$). For ENG we see the same effect for layer 1.

\paragraph{Other bias factors}
We can see in Table \ref{tbl: ClusterInfluence short} that most influences are above-baseline. As explained above, the word form bias heavily decreases using higher layers of BERT. For all other influences the bias increases when using high layers of BERT. This may be because decreasing the word form influence reveals the existence of further --less strong but still relevant-- influences. The same is observable with the Lemma and TokLem results, since there the form influence is decreased or even eliminated. While for ENG the influence scores mostly increase using Lemma and TokLem, for GER only the position influence increases, while corpora influence decreases. This is probably because the corpora influence is to some extent related to word form, which often reflects time-specific orthography as in \textit{Ackergeräth} vs. \textit{Ackergerät}, where the spelling with the "h" mostly occurs in the old corpus.  

Influence of position and proper names seems to be less important but the respective scores are still most of the times higher than the baselines. So overall the reflection of the two corpora seems to be the most influential factor apart from word form. Often the corpus bias is almost as high as the actual ARI score.

\subsection{Average Measures}
For the average measures we perform experiments for all three languages (ENG, GER, SWE).

\paragraph{Layers}
Because we observe a strong variation of influence scores with layers, as seen in Section (\ref{clustering}), we test different layer combinations for the average measures. The following are considered: 1, 12, 1+12, 1+2+3+4 (1-4), 9+10+11+12 (9-12). As shown in Table \ref{tab:performance1}, the choice of the layers strongly affects the performance. We see that for APD the higher layer combinations 12 and 9-12 perform best across all three languages, while the latter is slightly better ($.571$, $.407$ and $.554$). Interestingly, these two are the only layer combinations that do not include layer 1. All three layer combinations that include layer 1 are significantly worse in comparison. 
While COS performs best with layer combination 1-4 for ENG ($.390$), for GER and SWE we see a similar trend as with APD. Again, the higher layer combinations perform better than the other three, which all include layer 1. For GER layer combination 12 ($.472$) performs best, while 9-12 yields the highest result for SWE ($.183$). Our results are mostly in line with the findings of \citet{kutuzov-giulianelli-2020-uiouva} that APD works best on ENG and SWE, while COS yields the best scores for GER.

\begin{table}[t]
\center
\begin{adjustbox}{width=0.48\textwidth}
\begin{tabular}{ l | c c c | c c c }
\toprule
\multirow{2}{*}{\textbf{Layer}}& \multicolumn{3}{c |}{\textbf{APD}} & \multicolumn{3}{c}{\textbf{COS}} \\
& ENG & GER & SWE & ENG & GER & SWE \\
\hline
1 & .297 & .205 & .228 & .246 & .246 & .089 \\
12 & .566 & .359 & .529 & .339 & \textbf{.472} & .134 \\
1+12 & .455 & .316 & .280 & .365 & .373 & .077 \\
1-4 & .431 & .227 & .355 & \textbf{.390} & .297 & .079 \\
9-12 & \textbf{.571} & \textbf{.407} & \textbf{.554} & .365 & .446 & \textbf{.183} \\
\bottomrule
\end{tabular}
\end{adjustbox}
\caption{Token performance for different layer combinations across languages.}\label{tab:performance1}
\vspace{-7ex}
\end{table}

\paragraph{Pre-processing}
As with the clustering, we try to improve the performance of the average measures by using the two above-described pre-processing approaches. We perform experiments only for three layer combinations in order to reduce the complexity: (i) 12 and (ii) 9-12 perform best and are therefore obvious choices. (iii) From the remaining combinations 1+12 shows the most stable performance across measures and languages. Table \ref{tab:pre-process} shows the performance of the pre-processings (Lemma, TokLem) over these three combinations. We can see that both APD and COS perform slightly worse for ENG when paired with a pre-processing (exception to this is 1+12 Lemma). In contrast, GER profits heavily: While APD with layer combinations 12 and 9-12 performs slightly worse with Lemma, and slightly better with TokLem, we observe an enormous performance boost for layer combination 1+12 ($.643$ Lemma and $.731$ TokLem). We achieve a similar boost for all three layer combinations with COS as a measure. We reach a top performance of $.755$ for layer 12 with TokLem. SWE does not benefit from Lemma. We observe large performance decreases, with the exception of combination 1+12 (APD). The APD performance of layers 12 and 9-12 is slightly worse with TokLem. However, layers 1+12, which performed poorly without pre-processing, reaches peak performance of $.602$ with TokLem. All COS performances increase with TokLem, but are still well below the APD counterparts. The general picture is that GER and SWE profit strongly from TokLem. 

\begin{table}[t]
\centering
\begin{adjustbox}{width=0.43\textwidth}
\begin{tabular}{ c | c | l | c c c }
\toprule
& & \textbf{Layer} & \textbf{Token} & \textbf{Lemma} & \textbf{TokLem} \\
\hline
\multirow{6}{*}{\rotatebox[origin=c]{90}{\textbf{ENG}}} & \multirow{3}{*}{\rotatebox[origin=c]{90}{\textbf{APD}}} & 12 & \textbf{.566} & .483 & .494\\
& & 1+12 & .455 & \textbf{.483} & .455 \\
& & 9-12 & \textbf{.571} & .493 & .547 \\
\cline{2-6}
&\multirow{3}{*}{\rotatebox[origin=c]{90}{\textbf{COS}}} & 12 & \textbf{.339} & .251 & .331 \\
& & 1+12 & \textbf{.365} & .239 & .193 \\
& & 9-12 & \textbf{.365} & .286 & .353 \\
\midrule
\multirow{6}{*}{\rotatebox[origin=c]{90}{\textbf{GER}}} & \multirow{3}{*}{\rotatebox[origin=c]{90}{\textbf{APD}}} & 12 & .359 & .303 & \textbf{.456}\\
& & 1+12 & .316 & .643 & \textbf{.731} \\
& & 9-12 & .407 & .305 & \textbf{.516} \\
\cline{2-6}
&\multirow{3}{*}{\rotatebox[origin=c]{90}{\textbf{COS}}} & 12 & .472 & .693 & \textbf{.755} \\
& & 1+12 & .373 & .698 & \textbf{.729} \\
& & 9-12 & .446 & .689 & \textbf{.726} \\
\midrule
\multirow{6}{*}{\rotatebox[origin=c]{90}{\textbf{SWE}}} & \multirow{3}{*}{\rotatebox[origin=c]{90}{\textbf{APD}}} & 12 & \textbf{.529} & .214 & .505\\
& & 1+12 & .280 & .368 & \textbf{.602} \\
& & 9-12 & \textbf{.554} & .218 & .531 \\
\cline{2-6}
&\multirow{3}{*}{\rotatebox[origin=c]{90}{\textbf{COS}}} & 12 & .134 & -.019 & \textbf{.285} \\
& & 1+12 & .077 & .012 & \textbf{.082} \\
& & 9-12 & .183 & -.002 & \textbf{.284} \\
\bottomrule
\end{tabular}
\end{adjustbox}
\caption{Performance of pre-processing variants for three layer combinations.}
\vspace{-12mm}
\label{tab:pre-process}
\end{table}

\paragraph{Word form bias}
In order to better understand the effects of layer combinations and pre-processing, we compute correlations between word form and model predictions. To lessen the complexity, only layer combination 1+12 (which performed worst overall and includes layer 1), layer combination 9-12 (which performed best overall) in combination with Token and the superior TokLem are considered. The results are presented in Table \ref{tab:avm_wordform}. We observe similar findings for all three languages. The correlation between word form and APD predictions is strong ($.613$, $.554$ and $.730$) for layers 1+12 without pre-processing. The correlation is much weaker with layers 9-12 ($.068$, $.292$ and $.237$) or TokLem ($-.026$, $.105$ and $.176$). This is in line with the
performance development that also increases using layers 9-12 or TokLem. Both approaches (different layers, pre-processing) result in a considerable performance increase as described previously. Using layer combination 9-12 with TokLem further decreases the correlation (with the exception of ENG). However, the performance is better when only one of these approaches is used. 
The correlation between word form and COS model predictions is weaker overall ($.246$, $.387$ and $.429$). We see a similar correlation development as for APD, however this time the performance of ENG does not profit from the lowered bias (see Table \ref{tab:pre-process}). Both GER and SWE see a performance increase when the word form bias is lowered by either using layers 9-12 or TokLem. 

\begin{table}[]
\centering
\begin{adjustbox}{width=0.34\textwidth}
\begin{tabular}{c | c | l | c c }
\toprule
& & \textbf{Layer} & \textbf{Token} & \textbf{TokLem} \\
\hline
\multirow{4}{*}{\rotatebox[origin=c]{90}{\textbf{ENG}}} & \multirow{2}{*}{\rotatebox[origin=c]{90}{\textbf{APD}}} & 1+12 & .613 & -.026 \\
& & 9-12 & .068 & .090 \\
\cline{2-5}
& \multirow{2}{*}{\rotatebox[origin=c]{90}{\textbf{COS}}} & 1+12 & .246 &  -.062 \\
& & 9-12 & .020 &  .004 \\
\midrule
\multirow{4}{*}{\rotatebox[origin=c]{90}{\textbf{GER}}} & \multirow{2}{*}{\rotatebox[origin=c]{90}{\textbf{APD}}} & 1+12 & .554 & .271 \\
& & 9-12 & .292 & .105 \\
\cline{2-5}
& \multirow{2}{*}{\rotatebox[origin=c]{90}{\textbf{COS}}} & 1+12 & .387 &  -.017 \\
& & 9-12 & .205 &  -.008 \\
\midrule
\multirow{4}{*}{\rotatebox[origin=c]{90}{\textbf{SWE}}} & \multirow{2}{*}{\rotatebox[origin=c]{90}{\textbf{APD}}} & 1+12 & .730 & .176 \\
& & 9-12 & .237 & .048 \\
\cline{2-5}
& \multirow{2}{*}{\rotatebox[origin=c]{90}{\textbf{COS}}} & 1+12 & .429 &  -.031 \\
& & 9-12 & .277 & -.035 \\
\bottomrule
\end{tabular}
\end{adjustbox}
\caption{Correlations of word form and predicted change scores.}
\vspace{-12mm}
\label{tab:avm_wordform}
\end{table}

\paragraph{Polysemy bias}
The SemEval data sets are strongly biased by polysemy, i.e., a perfect model measuring the true synchronic target word polysemy in either $t_1$ or $t_2$ could reach above $.7$ performance \citep{schlechtweg-etal-2020-semeval}. We use APD-OLD and APD-NEW (see Section \ref{sec:models}) to see whether we can exploit this fact to create a purely synchronic polysemy model with high performance. We achieve moderate performances for ENG and GER ($.274$/$.332$ and $.321$/$.450$ respectively) and a good performance for SWE ($.550$/$.562$). While the performance for ENG and GER is clearly below the high-scores, the performance is high for a measure that lacks any kind of diachronic information. And in the case of SWE, the performance of both APD-OLD and APD-NEW is just barely below the high-scores (cf. Table \ref{tab:pre-process}). Note that regular APD (in contrast to COS) is, in theory, affected by polysemy \citep{Schlechtwegetal18}. It is thus possible that APD's high performance stems at least partly from this polysemy bias. This is supported by comparing the SWE results of APD and COS in Table \ref{tab:pre-process}: COS is weakly influenced by polysemy and performs poorly, while APD has higher performance, but only slightly above the purely synchronic measures APD-OLD/NEW.

\section{Conclusion}
BERT token representations are influenced by various factors, but most strongly by target word form. Even in higher layers this influence persists. By removing the form bias we were able to considerably improve the performance across languages. Although we reach comparably high performance with clustering for graded change detection in German, average measures still perform better than cluster-based approaches. The reasons for this are still unclear and should be addressed in future research. Furthermore, we used BERT without fine-tuning. It would be interesting to see how fine-tuning interacts with influence variables and whether it further improves performance. 

\bibliography{Bibliography-general,bibliography-self,additional_references}
\bibliographystyle{acl_natbib}

\newpage
\appendix

\section{Word Usage Graphs}
\label{sec:wugs}
\begin{figure*}[h!]
    \begin{subfigure}{.33\textwidth}
\frame {        \includegraphics[width=\linewidth]{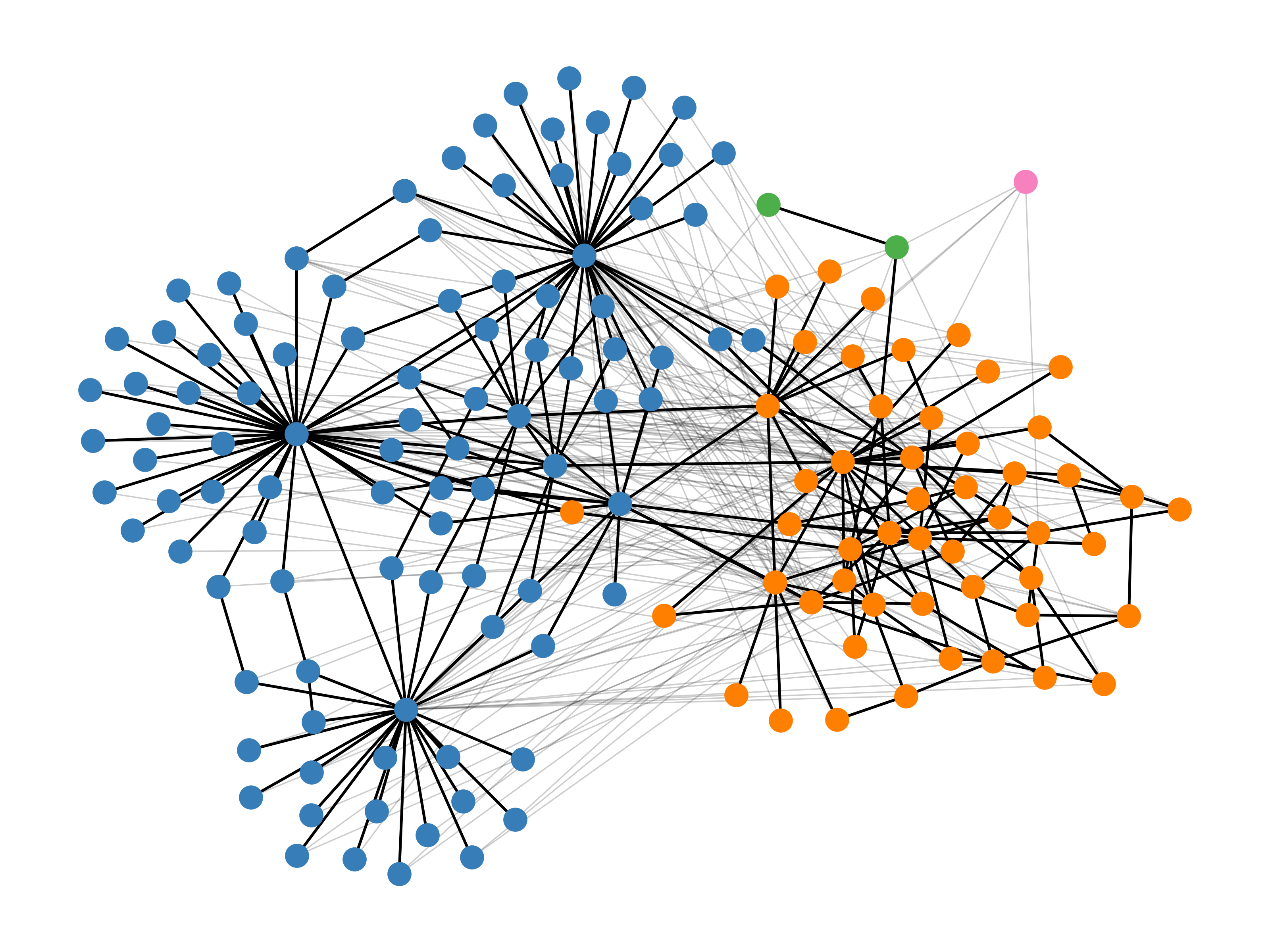}}
        \caption*{full}
    \end{subfigure}
    \begin{subfigure}{.33\textwidth}
\frame{        \includegraphics[width=\linewidth]{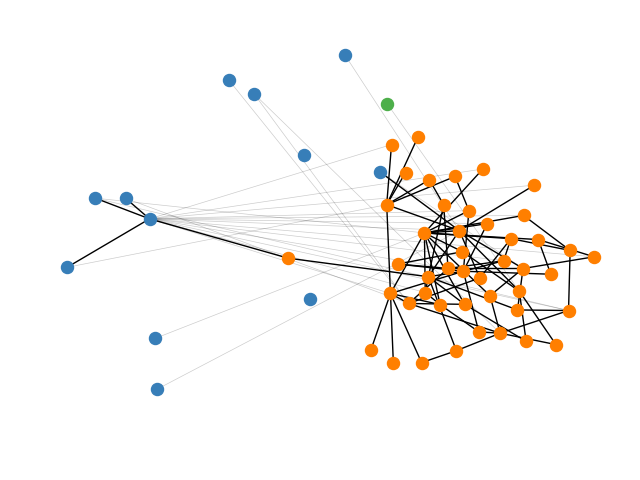}}
        \caption*{$t_1$}
    \end{subfigure}%
    \begin{subfigure}{.33\textwidth}
\frame{        \includegraphics[width=\linewidth]{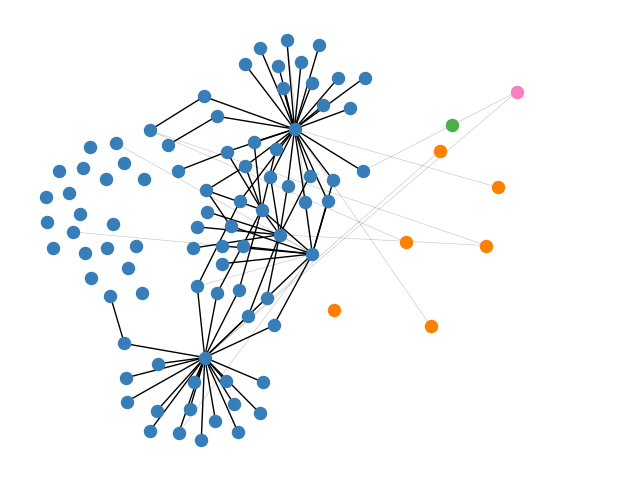}}
        \caption*{$t_2$}
    \end{subfigure}%
    \caption{Word Usage Graph of German \textit{Eintagsfliege}.
    Nodes represent uses of the target word. Edge weights represent the median of relatedness judgments between uses (\textbf{black}/\textcolor{gray}{gray} lines for \textbf{high}/\textcolor{gray}{low} edge weights). Colors indicate clusters (senses) inferred from the full graph. $D_1=( 12, 45, 0, 1)$, $D_2=( 85, 6, 1, 1)$, $B(w)=0$ and $G(w)=0.66$.}\label{fig:graph1}
\end{figure*}

Please find an example of a Word Usage Graph (WUG) for the German word \textit{Eintagsfliege} in Figure \ref{fig:graph1} \citep{schlechtweg-etal-2020-semeval,Schlechtweg2021dwug}.

\begin{table*}[h!]
\center
\begin{adjustbox}{width=0.42\textwidth}
\begin{tabular}{ c | c | l | c c c }
\toprule
& & \textbf{Layer} & \textbf{Token} & \textbf{Lemma} & \textbf{TokLem} \\
\midrule
\multirow{10}{*}{\rotatebox[origin=c]{90}{\textbf{Performance}}} &

\multirow{5}{*}{\rotatebox[origin=c]{90}{\textbf{$\rho$}}}
& 1 & -.141 & -.033 & .100\\
& & 12 & .205 & .154 & .168 \\
& & 1+12 & -.316 & .130 & .081 \\
& & 6+7 & .075 & -.103 & .017\\
& & 9-12 & .325 & \textbf{.345} & .293 \\

\cline{2-6}
&\multirow{5}{*}{\rotatebox[origin=c]{90}{\textbf{ARI}}}
& 1 & .022 & .041 & .045\\
& & 12 & .116 & .111 & .158 \\
& & 1+12 & .022 & .141 & .149 \\
& & 6+7 & .119 & .111 & .145\\
& & 9-12 & .150 & .159 & \textbf{.163} \\

\bottomrule
\end{tabular}
\end{adjustbox}
\quad
\begin{adjustbox}{width=0.42\textwidth}
\begin{tabular}{ c | c | l | c c c }
\toprule
& & \textbf{Layer} & \textbf{Token} & \textbf{Lemma} & \textbf{TokLem} \\
\midrule

\multirow{10}{*}{\rotatebox[origin=c]{90}{\textbf{Performance}}} &

\multirow{5}{*}{\rotatebox[origin=c]{90}{\textbf{$\rho$}}}
& 1              &      -.265      &      -.062 &      -.170\\
& & 12           &      .123      &      .427 &      \textbf{.624}\\
& & 1+12         &      -.252      &      .235 &      .401\\
& &   6+7        &      .002     &      .464 &      .320\\
& & 9-12   &      .122      &      .420 &      .533\\

\cline{2-6}
&\multirow{5}{*}{\rotatebox[origin=c]{90}{\textbf{ARI}}}
& 1              &      .033      &      .002 &      .003\\
& & 12           &      .119      &      .159 &      \textbf{.161}\\
& & 1+12         &      .037      &      .064 &      .080\\
& &    6+7       &      .101      &      .158 &      .152\\
& & 9-12   &      .155      &      .142 &      .154\\
\bottomrule
\end{tabular}
\end{adjustbox}
\caption{English clustering performance (left) and German clustering performance (right).}
\label{tbl: Clusterperformance}
\end{table*}

\newpage

\section{Extended clustering performances and influences}
\label{sec:performance}

Please find the full results of our cluster experiments in Tables \ref{tbl: Clusterperformance} and \ref{tbl: ClusterInfluence}.

\begin{table*}[h!]
\center
\begin{adjustbox}{width=0.39\textwidth}
\begin{tabular}{ c | c | l | c c c }
\toprule
& & \textbf{Layer} & \textbf{Token} & \textbf{Lemma} & \textbf{TokLem} \\
\midrule
\multirow{15}{*}{\rotatebox[origin=c]{90}{\textbf{Form}}} &

\multirow{5}{*}{\rotatebox[origin=c]{90}{\textbf{Influence}}}
& 1 & .907 & .014 & .014\\
& & 12 & .389 & .018 & .077 \\
& & 1+12 & .881 & .020 & .057 \\
& & 6+7 & .572 & .013 & .028\\
& & 9-12 & .334 & .018 & .051 \\

\cline{2-6}
&\multirow{5}{*}{\rotatebox[origin=c]{90}{\textbf{Random}}}
& 1 & .002 & .002 & .002\\
& & 12 & -.001 & .001 & -.001 \\
& & 1+12 & -.002 & -.001 & -.001 \\
& & 6+7 & .001 & .002 & .001\\
& & 9-12 & -.001 & -.001 & -.002 \\

\cline{2-6}
&\multirow{5}{*}{\rotatebox[origin=c]{90}{\textbf{Baseline}}}
& 1 & .017 & .017 & .017\\
& & 12 & .017 & .017 & .017 \\
& & 1+12 & .017 & .017 & .017 \\
& & 6+7 & .017 & .017 & .017\\
& & 9-12 & .017 & .017 & .017 \\
\midrule
\multirow{15}{*}{\rotatebox[origin=c]{90}{\textbf{Position}}} &

\multirow{5}{*}{\rotatebox[origin=c]{90}{\textbf{Influence}}}
& 1 & .001 & .026 & .024\\
& & 12 & .012 & .012 & .015 \\
& & 1+12 & -.001 & .019 & .007 \\
& & 6+7 & -.002 & .018 & -.003\\
& & 9-12 & .002 & .007 & .003 \\

\cline{2-6}
&\multirow{5}{*}{\rotatebox[origin=c]{90}{\textbf{Random}}}
& 1 & .001 & .003 & .001\\
& & 12 & .001 & -.001 & -.001 \\
& & 1+12 & -.001 & -.001 & -.001 \\
& & 6+7 & .001 & -.001 & -.001\\
& & 9-12 & .001 & .001 & -.001 \\

\cline{2-6}
&\multirow{5}{*}{\rotatebox[origin=c]{90}{\textbf{Baseline}}}
& 1 & -.002 & -.002 & -.002\\
& & 12 & -.002 & -.002 & -.002 \\
& & 1+12 & -.002 & -.002 & -.002 \\
& & 6+7 & -.002 & -.002 & -.002\\
& & 9-12 & -.002 & -.002 & -.002 \\
\midrule
\multirow{15}{*}{\rotatebox[origin=c]{90}{\textbf{Corpora}}} &

\multirow{5}{*}{\rotatebox[origin=c]{90}{\textbf{Influence}}}
& 1 & .019 & .021 & .033\\
& & 12 & .078 & .056 & .082 \\
& & 1+12 & .027 & .050 & .074 \\
& & 6+7 & .034 & .035 & .050\\
& & 9-12 & .056 & .044 & .063 \\

\cline{2-6}
&\multirow{5}{*}{\rotatebox[origin=c]{90}{\textbf{Random}}}
& 1 & .001 & -.001 & .003\\
& & 12 & .001 & .001 & .001\\
& & 1+12 & -.001 & .001 & .001 \\
& & 6+7 & .001 & .001 & .002\\
& & 9-12 & .002 & .001 & .002 \\

\cline{2-6}
&\multirow{5}{*}{\rotatebox[origin=c]{90}{\textbf{Baseline}}}
& 1 & .018 & .018 & .018\\
& & 12 & .018 & .018 & .018 \\
& & 1+12 & .018 & .018 & .018 \\
& & 6+7 & .018 & .018 & .018\\
& & 9-12 & .018 & .018 & .018 \\
\midrule
\multirow{15}{*}{\rotatebox[origin=c]{90}{\textbf{Names}}} &

\multirow{5}{*}{\rotatebox[origin=c]{90}{\textbf{Influence}}}
& 1 & -.007 & .010 & .010\\
& & 12 & .025 & .027 & .033 \\
& & 1+12 & .018 & .022 & .027 \\
& & 6+7 & .012 & .016 & .027\\
& & 9-12 & .019 & .022 & .026 \\

\cline{2-6}
&\multirow{5}{*}{\rotatebox[origin=c]{90}{\textbf{Random}}}
& 1 & -.001 & -.002 & -.002\\
& & 12 & -.001 & .001 & .001 \\
& & 1+12 & -.001 & .001 & -.001 \\
& & 6+7 & -.001 & .001 & .001\\
& & 9-12 & -.001 & -.001 & .001 \\

\cline{2-6}
&\multirow{5}{*}{\rotatebox[origin=c]{90}{\textbf{Baseline}}}
& 1 & .019 & .019 & .019\\
& & 12 & .019 & .019 & .019 \\
& & 1+12 & .019 & .019 & .019 \\
& & 6+7 & .019 & .019 & .019\\
& & 9-12 & .019 & .019 & .019 \\
\bottomrule
\end{tabular}
\end{adjustbox}
\quad
\begin{adjustbox}{width=0.39\textwidth}
\begin{tabular}{ c | c | l | c c c }
\toprule
& & \textbf{Layer} & \textbf{Token} & \textbf{Lemma} & \textbf{TokLem} \\
\midrule

\multirow{15}{*}{\rotatebox[origin=c]{90}{\textbf{Form}}} &

\multirow{5}{*}{\rotatebox[origin=c]{90}{\textbf{Influence}}}
& 1              &      .706      &      .024 &      .004\\
& & 12           &      .439     &      .056 &      .150\\
& & 1+12         &      .687      &      .039 &      .046\\
& &    6+7       &      .503      &      .050 &      .050\\
& & 9-12   &      .420      &      .047 &      .094\\

\cline{2-6}
&\multirow{5}{*}{\rotatebox[origin=c]{90}{\textbf{Random}}}
& 1              &      -.001      &      -.002 &      .020\\
& & 12           &      -.001      &      .001 &      .021\\
& & 1+12         &      -.001      &      -.001 &      .020\\
& &  6+7         &      .002      &      .001 &      .019\\
& & 9-12   &      .001      &      -.001 &      .021\\

\cline{2-6}
&\multirow{5}{*}{\rotatebox[origin=c]{90}{\textbf{Baseline}}}
& 1              &      .036      &      .036 &      .036\\
& & 12           &      .036      &      .036 &      .036\\
& & 1+12         &      .036      &      .036 &      .036\\
& &      6+7     &      .036      &      .036 &      .036\\
& & 9-12   &      .036      &      .036 &      .036\\
\midrule
\multirow{15}{*}{\rotatebox[origin=c]{90}{\textbf{Position}}} &

\multirow{5}{*}{\rotatebox[origin=c]{90}{\textbf{Influence}}}
& 1              &      .005      &      .023 &      .027\\
& & 12           &      -.002      &      .005 &      -.002\\
& & 1+12         &      .002      &      .021 &      .013\\
& &   6+7        &      .010      &      .020 &      .018\\
& & 9-12   &      .009      &      .018 &      .012\\

\cline{2-6}
&\multirow{5}{*}{\rotatebox[origin=c]{90}{\textbf{Random}}}
& 1              &      .001      &      .001 &      .001\\
& & 12           &      .001      &      -.001 &      .001\\
& & 1+12         &      -.001      &      -.001 &      .002\\
& &   6+7        &      -.001      &      .001 &      .001\\
& & 9-12   &      -.001      &      .001 &      .001\\

\cline{2-6}
&\multirow{5}{*}{\rotatebox[origin=c]{90}{\textbf{Baseline}}}
& 1              &      .005      &      .005 &      .005\\
& & 12           &      .005      &      .005 &      .005\\
& & 1+12         &      .005      &      .005 &      .005\\
& &  6+7         &      .005      &      .005 &      .005\\
& & 9-12   &      .005      &      .005 &      .005\\
\midrule
\multirow{15}{*}{\rotatebox[origin=c]{90}{\textbf{Corpora}}} &

\multirow{5}{*}{\rotatebox[origin=c]{90}{\textbf{Influence}}}
& 1              &      .074      &      .003 &      .005\\
& & 12           &      .110      &      .095 &      .096\\
& & 1+12         &      .077      &      .024 &      .052\\
& &   6+7        &      .101      &      .058 &      .075\\
& & 9-12   &      .107     &      .068 &      .089\\

\cline{2-6}
&\multirow{5}{*}{\rotatebox[origin=c]{90}{\textbf{Random}}}
& 1              &      -.001      &      -.001 &      .001\\
& & 12           &      .001      &      -.001 &      .001\\
& & 1+12         &      -.001      &      .001 &      .002\\
& &   6+7        &      -.001      &      .001 &      -.001\\
& & 9-12   &      -.001      &      .001 &      -.001\\

\cline{2-6}
&\multirow{5}{*}{\rotatebox[origin=c]{90}{\textbf{Baseline}}}
& 1              &      .083      &      .083 &      .083\\
& & 12           &      .083      &      .083 &      .083\\
& & 1+12         &      .083      &      .083 &      .083\\
& &   6+7        &      .083      &      .083 &      .083\\
& & 9-12   &      .083      &      .083 &      .083\\
\midrule
\multirow{15}{*}{\rotatebox[origin=c]{90}{\textbf{Names}}} &

\multirow{5}{*}{\rotatebox[origin=c]{90}{\textbf{Influence}}}
& 1              &      -      &      - &      -\\
& & 12           &      -      &      - &      -\\
& & 1+12         &      -      &      - &      -\\
& &    6+7       &      -      &      - &      -\\
& & 9-12   &      -      &      - &      -\\

\cline{2-6}
&\multirow{5}{*}{\rotatebox[origin=c]{90}{\textbf{Random}}}
& 1              &      -      &      - &      -\\
& & 12           &      -      &      - &      -\\
& & 1+12         &      -      &      - &      -\\
& & 6+7          &      -      &      - &      -\\
& & 9-12   &      -      &      - &      -\\

\cline{2-6}
&\multirow{5}{*}{\rotatebox[origin=c]{90}{\textbf{Baseline}}}
& 1              &      -      &      - &      -\\
& & 12           &      -      &      - &      -\\
& & 1+12         &      -      &      - &      -\\
& &  6+7         &      -      &      - &      -\\
& & 9-12   &      -      &      - &      -\\

\bottomrule
\end{tabular}
\end{adjustbox}
\caption{English clustering influences (left) and German clustering influences (right).}
\label{tbl: ClusterInfluence}
\end{table*}

\end{document}